\documentclass[10pt,twocolumn,letterpaper]{article}

\usepackage{cvpr}
\usepackage{times}
\usepackage{epsfig}
\usepackage{graphicx}
\usepackage{amsmath}
\usepackage{amssymb}
\usepackage{enumitem}
\usepackage{floatrow}
\usepackage[font=small]{subfig}

\usepackage[pagebackref=true,breaklinks=true,letterpaper=true,colorlinks,bookmarks=false]{hyperref}

\cvprfinalcopy 

\def\cvprPaperID{3391} 

\begin{document}

\title{DDLSTM: Dual-Domain LSTM for Cross-Dataset Action Recognition} 

\author{Toby Perrett and Dima Damen\\
Department of Computer Science, University of Bristol, Bristol, UK\\
{\tt\small <firstname>.<lastname>@bristol.ac.uk}
}

\maketitle

\begin{abstract}
Domain alignment in convolutional networks aims to learn the degree of layer-specific feature alignment beneficial to the joint learning of source and target datasets. While increasingly popular in convolutional networks, there have been no previous attempts to achieve domain alignment in recurrent networks. Similar to spatial features, both source and target domains are likely to exhibit temporal dependencies that can be jointly learnt and aligned.

In this paper we introduce Dual-Domain LSTM (DDLSTM), an architecture that is able to learn temporal dependencies from two domains concurrently.  It performs cross-contaminated batch normalisation on both input-to-hidden and hidden-to-hidden weights, and learns the parameters for cross-contamination, for both single-layer and multi-layer LSTM architectures. We evaluate DDLSTM on frame-level action recognition using three datasets, taking a pair at a time, and report an average increase in accuracy of~3.5\%. The proposed DDLSTM architecture outperforms standard, fine-tuned, and batch-normalised LSTMs.
\end{abstract}

\vspace{-4mm}

\section{Introduction}
Online action recognition has direct implications on assistive and surveillance applications, enabling action classification as soon as a new frame is observed.  It only depends on previously observed frames, with no knowledge from future observations.
This contrasts offline action recognition where
the whole action is observed before being classified.

One obstacle to deploying online action-recognition systems in the wild is that they require a large amount of training data to achieve high performance, so an open question is how to make best-use of multiple datasets to achieve higher recognition accuracy. 
{Such cross-dataset temporal dependencies can be present even when datasets use different class labels.}
In this paper, we focus on the related tasks of kitchen activities. In one dataset, a sequence of actions could be labeled as:\\ \hspace*{6pt}\textcolor{blue}{`pick-up knife' $\rightarrow$ `cut onion' $\rightarrow$ `put onion in pan'}\\ whereas a second dataset would have labels such as:\\ \hspace*{6pt}\textcolor{blue}{`take knife' $\rightarrow$ `chop potato' $\rightarrow$ `place on a baking tray'}\\ While the two sets of labels differ, we investigate how a joint recurrent model can be learnt for both datasets and we demonstrate that this joint training outperforms independently-learnt models. 
\vspace{-0mm}

In online recognition, recurrent models are typically used, particularly Long Short-Term Memory recurrent networks (LSTM)~\cite{Du2015,Donahue2017,Liu2018}.
It has recently been shown that 
LSTMs can benefit from multiple data sources when using CNN features for frame-based action classification \cite{Perrett}. 
However, LSTMs are not explicitly designed to handle information from multiple domains.  We aim to address this limitation by combining recent advances in CNN domain adaptation~\cite{Carlucci2017} with batch-normalised LSTM training \cite{Cooijmans2016}, and introduce the Dual-Domain LSTM (DDLSTM).
We show that it is indeed possible for the DDLSTM to learn jointly from two related datasets, and in a way which outperforms the standard LSTM for frame-based online action recognition (whether jointly trained, or pre-trained and fine-tuned).
{Importantly, our formalism allows learning cross-contamination parameters in a differentiable manner within the backpropation-through-time of the LSTM.} 
In each experiment, we evaluate DDLSTM on pairs of datasets, out of three datasets frequently used for kitchen-based activities: 50 Salads \cite{Stein}, Breakfast \cite{Kuehne2014} and MPII Cooking 2 \cite{Rohrbach2012}. We also demonstrate the benefit of joint training with larger datasets of related domains (e.g. EPIC~\cite{Damen2018EPICKITCHENS}) to leverage missing temporal knowledge.

The rest of this paper is organised as follows.  Section \ref{ch:bg} gives a summary of related domain adaptation and LSTM literature.  Section \ref{ch:method} introduces the proposed DDLSTM.  Section \ref{ch:data} gives an overview of the datasets we use for evaluation.  Section \ref{ch:exp} includes the comparative analysis, where we show an average increase in frame-level recognition accuracy of 3.5\%.  Finally, the conclusion is in Section \ref{ch:conclusion}.

\section{Background}\label{ch:bg}

Works which learn from multiple domains have traditionally used measures, such as the Maximum Mean Discrepancy (MMD) \cite{Gretton2006,Long2013}, to determine differences in feature space between the domains, and apply transformations to bring them closer together~\cite{Busto2017,B2017}.
More recently, loss functions which take these differences into account have been used, in conjunction with classification loss, in CNNs to merge domains in an end to end fashion \cite{Haeusser2017,Csurka2017,Venkateswara2017},  enabling predictions to be made on unsupervised domains.
Recent advances have taken the approach of introducing additional layers into a network in order to align deep and shallow feature spaces.  Of particular interest to us is the work of 
Carlucci et al. \cite{Carlucci2017}.  Their Automatic Domain Alignment Layers are based around expanding batch normalisation (introduced by Ioffe and Szegedy \cite{Ioffe2015} to improve model accuracy and reduce the number of training iterations) to handle inputs from source and target domains.  This is achieved by calculating separate batch statistics from source and target samples within a batch, but allowing for some cross-contamination between domains.  These layers are inserted into classification networks (after the fully connected layers in AlexNet \cite{Krizhevsky2012} and replacing standard batch normalisation layers in InceptionBN), equipped with a softmax classification loss for source samples and an entropy loss for unlabelled target samples.
{This approach is related to, but fundamentally different from cross-network stitching for multi-tasks~\cite{Misra2016}. In ~\cite{Misra2016}, two networks which perform different tasks (e.g. detection vs captioning) on \emph{the same input} are trained. However in~\cite{Carlucci2017}, one task (e.g. classification) is performed}
on \textit{two sets input features} which are not necessarily related.

The works covered so far are all designed for classification {using convolutional networks} trained on single images.  For example, a commonly used benchmark is the Office+Caltech dataset \cite{Gong2012} where adaptation between images taken with a DSLR in an office and product images taken from Amazon is attempted.  The most common technique to utilise multiple datasets in video-level classification is to train on one and fine tune on another ~\cite{Carreira}. Attempts have also been made to use semantic similarity between labels~\cite{Zheng2009}, and to treat different camera angles of the same content as different domains \cite{Nie}.

We are specifically interested in online action recognition, where single-dataset methods tend to use CNN features with LSTMs to make frame-level predictions \cite{Ng,Wu2014}.Incorporating a domain adaptation component into recurrent neural networks (RNNs) has the potential to provide a number of benefits.
Applying domain adaptation within an RNN will allow the direct use of different features which are well suited to their respective domains, whilst also enabling the learning of related temporal information from multiple domains at the same time.  Additionally, when RNNs are used with raw sensor output, the only place where domain adaptation can occur is within the RNN.

LSTMs are an obvious candidate for modification to become multi-domain aware, as their ability to remember information for a long period of time makes them particularly well suited to applications such as frame-based action recognition (which we focus on here), and language modelling, amongst others.
Greff et al. \cite{Greff2016} evaluated 8 different LSTM cell types, and found there was very little difference between them.  A number of works have attempted to incorporate batch normalisation into an LSTM framework \cite{Laurent2016,Cooijmans2016}.  In~\cite{Cooijmans2016}, Recurrent Batch Normalisation consists of normalisation of two sets of weights;  Input-to-hidden weight normalisation can be thought of as standard batch normalisation, and this is used in conjunction with hidden-to-hidden normalisation.  They found that performing a separate normalisation for each timestep achieved better performance due to initial activations being dissimilar to values which are converged to after multiple timesteps. We revisit the batch-normalised LSTM (BNLSTM) in Section~\ref{ch:method}.

\begin{figure*}
\centering
\subfloat[Standard LSTM trained on one dataset.\label{fig:lstm_train_1}]{\includegraphics[trim=-70 0 -70 0,clip,scale=0.25]{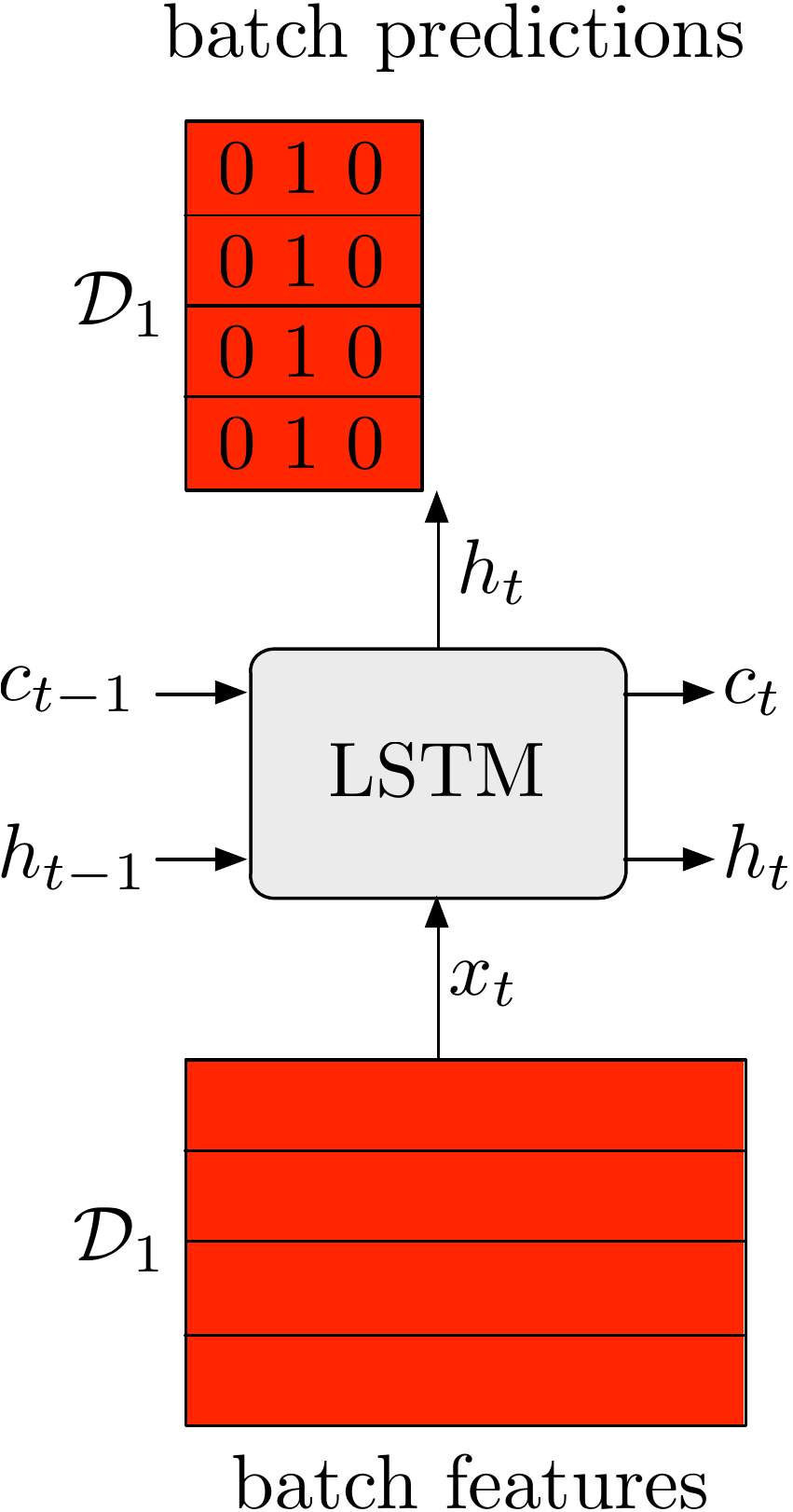}} \hspace{10pt}
\subfloat[LSTM trained jointly on two datasets, with concatenated labels.\label{fig:lstm_train_2}]{\includegraphics[trim=-70 0 -70 0,clip,scale=0.25]{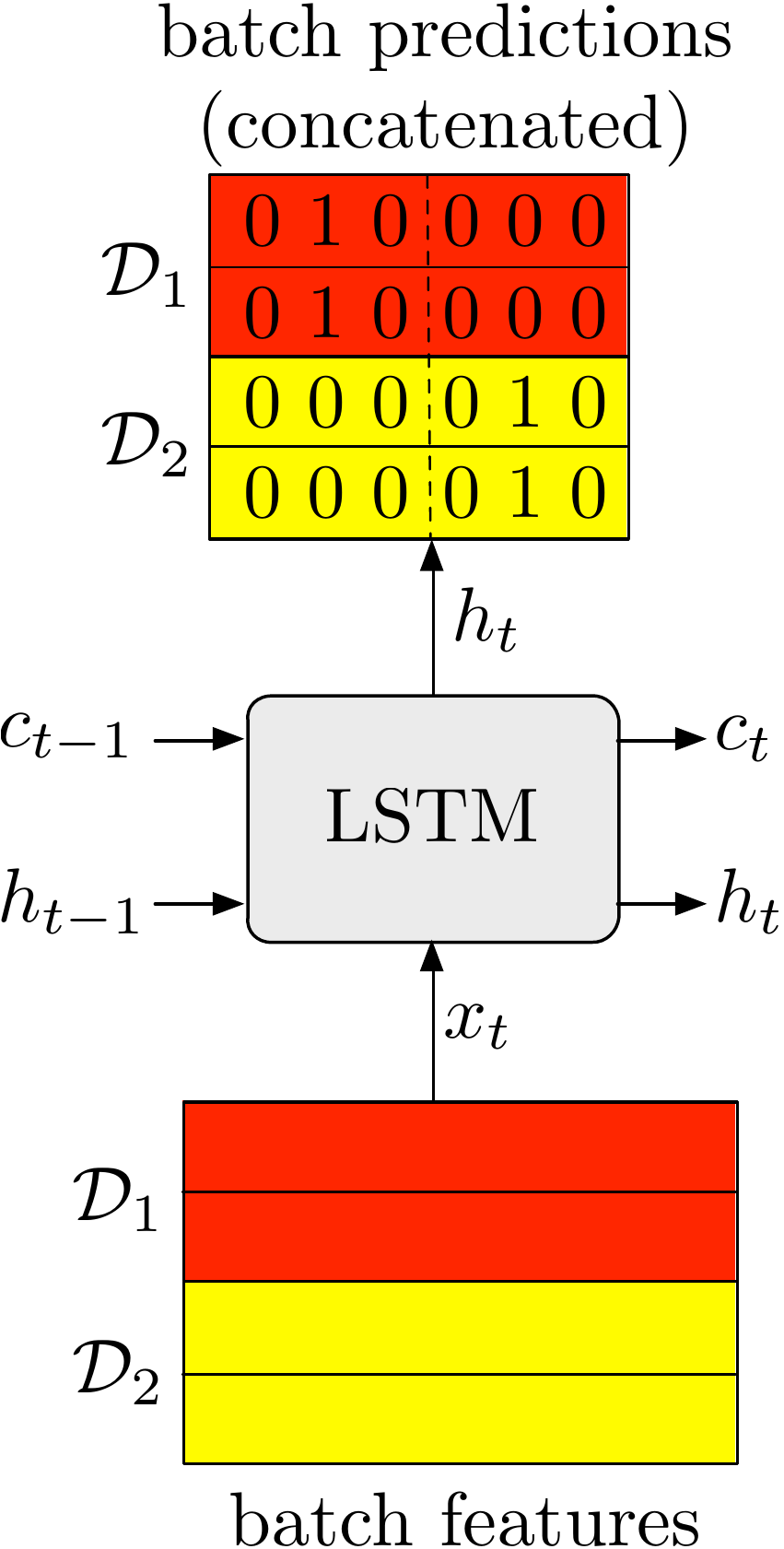}} 
\hspace{10pt}
\subfloat[BNLSTM trained on one dataset.  BN on feature input and hidden-to-hidden input and output.\label{fig:lstm_train_3}]{\includegraphics[scale=0.25]{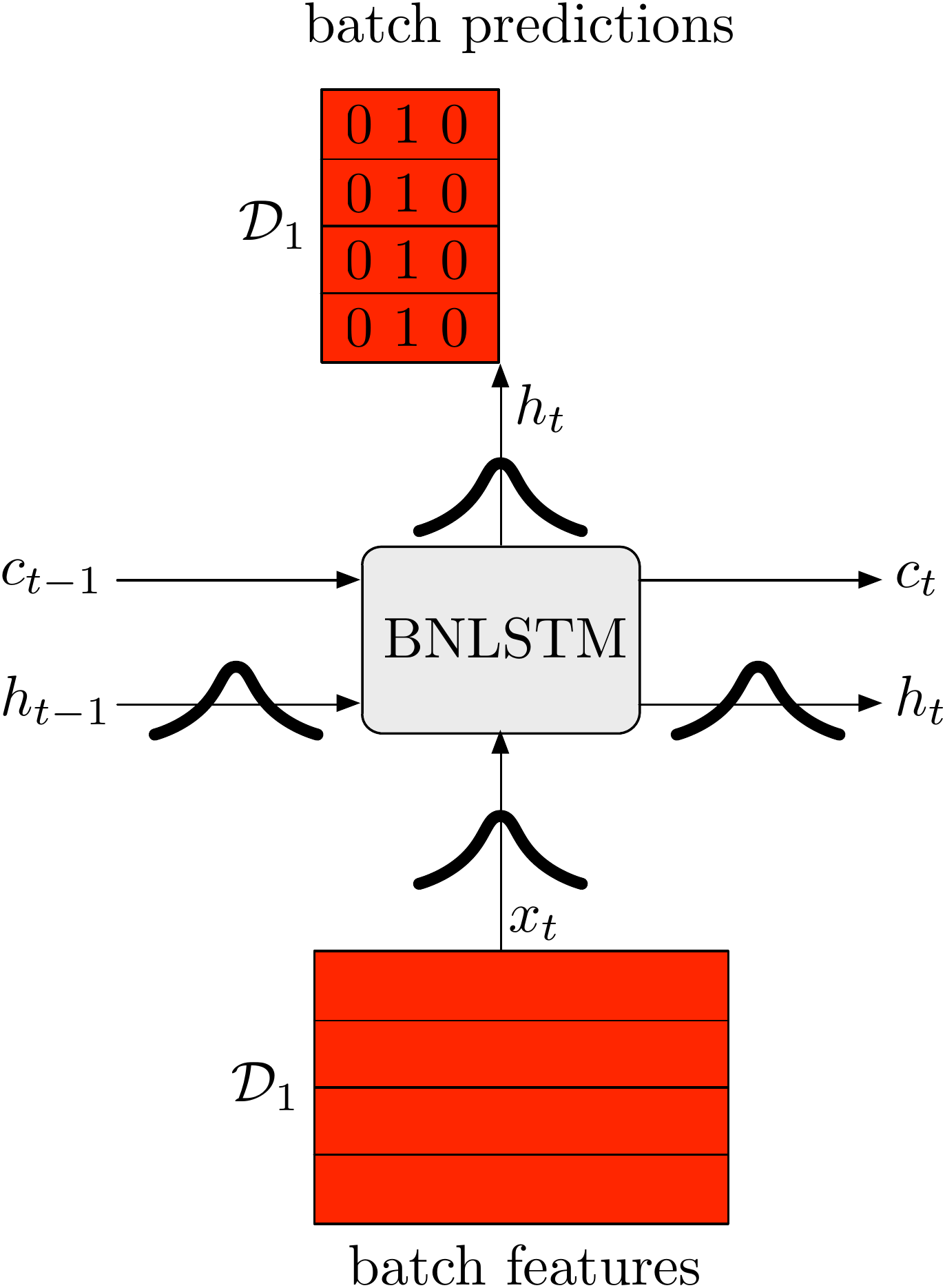}} \hspace{10pt}
\subfloat[DDLSTM with concatenated labels.  Dual-domain BN on feature and hidden-to-hidden inputs.\label{fig:lstm_train_4}]{\includegraphics[scale=0.25]{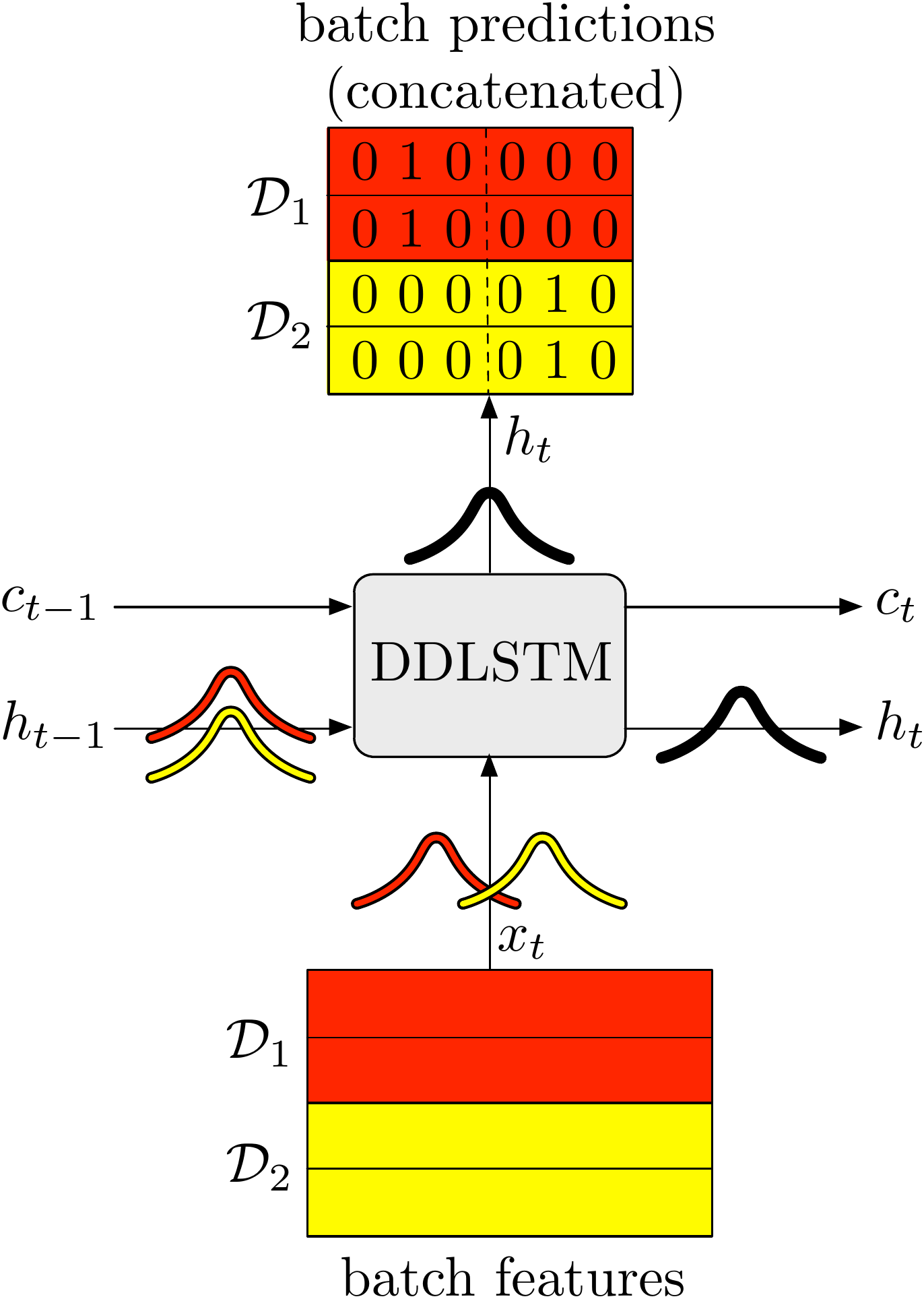}}
\vspace*{-6pt}
\caption{Proposed DDLSTM (d) in comparison to other LSTM architectures and training procedures: (a) single domain LSTM, (b) jointly trained LSTM, (c) single domain batch-normalised LSTM.}
\label{fig:comparison}
\end{figure*}

\section{Dual-Domain LSTM}\label{ch:method}

In this section, we propose the DDLSTM, capable of operating on arbitrary sequential data, which we evaluate in the context of online action recognition.  It is able to jointly learn temporal dependencies from two domains: both independent temporal dependencies per domain, as well as common cross-domain temporal dependencies.

Fig. \ref{fig:lstm_train_1} shows a standard LSTM trained on one dataset.  Fig. \ref{fig:lstm_train_2} shows a standard LSTM trained jointly on two datasets (i.e. each batch contains examples from both). This generalises to cases where the labels in the two datasets do not necessarily match. Should the labels match, a single shared output vector can be used. However, when each domain has its own set of labels, predictions can be defined as concatenated label vectors, with the first part of the output vector corresponding to predictions for the first dataset's labels, and the second part corresponding to predictions for the second dataset's labels. While this architecture learns the mapping from both input domains to shared or distinct output labels, the model is likely to learn each domain independently, as no effort to align the input from both domains is incorporated in the architecture.

Before introducing the DDLSTM for {aligning and jointly learning from two domains}, we revisit the BNLSTM for single domain batch normalisation from \cite{Cooijmans2016}.
Fig. \ref{fig:lstm_train_3} shows the BNLSTM trained on a single dataset.  
We choose to base the DDLSTM on the BNLSTM architecture for two main reasons.  First, the BNLSTM demonstrated superior results over the standard LSTM in applications such as language modelling and simple sequential MNIST \cite{Cooijmans2016}.  Second, and perhaps more importantly, it incorporates batch normalisation, which makes it suitable for adaptation to work with the domain-mixing aspect (via batch normalisation) of automatic domain alignment layers.  Its formulation in \cite{Cooijmans2016} is given as:
\begin{equation}\label{eq:bnlstm_f}
\left( \begin{array}{c}
\mathbf{\tilde f}_t \\
\mathbf{\tilde i}_t \\
\mathbf{\tilde o}_t \\
\mathbf{\tilde g}_t \\
\end{array} \right)
\begin{array}{ll}
 = & \text{BN} \left( \mathbf{W}_h\mathbf{h}_{t-1};\gamma_h,\beta_h \right) \\
  & + \text{BN} \left( \mathbf{W}_x\mathbf{x}_{t};\gamma_x,\beta_x \right)
 + \mathbf{b}
\end{array}
\end{equation}
\begin{equation}\label{eq:bnlstm_c}
\mathbf{c}_t = \sigma\left(\mathbf{\tilde f}_t\right) \odot \mathbf{c}_{t-1} + \sigma\left(\mathbf{\tilde i}_t\right) \odot \text{tanh}\left(\mathbf{\tilde g}_t\right)
\end{equation}
\begin{equation}\label{eq:bnlstm_h}
\mathbf{h}_t = \sigma\left(\mathbf{\tilde o}_t\right) \odot \text{tanh} \left( \text{BN}\left(\mathbf{c}_t;\gamma_c,\beta_c \right) \right)	
\end{equation}
Equation \ref{eq:bnlstm_f} contains the forget gate layer ($\mathbf{\tilde f}$),  input gate layer ($\mathbf{\tilde i}$), output gate layer~($\mathbf{\tilde o}$) and layer used to generate candidates to change the cell state later on~($\mathbf{\tilde g}$).
Here, the normalisation of the $\mathbf{W}_x\mathbf{x}_{t}$ term can be thought of as input-to-hidden normalisation (as it operates on the input at the current timestep, $\mathbf{x}_{t}$).  The normalisation of the $\mathbf{W}_h\mathbf{h}_{t-1}$ term can be thought of as hidden-to-hidden normalisation, as it operates on the output of the cell at the previous timestep~($\mathbf{h}_{t-1}$).
Equation \ref{eq:bnlstm_c} gives the new cell state $\mathbf{c}_t$.  Note how it is not normalised, which allows the gradient to flow through timesteps.
Equation \ref{eq:bnlstm_h} gives the cell output, where $\mathbf{c_t}$ is normalised to match the  $\mathbf{\tilde o}$ term.
The batch normalisation function \cite{Cooijmans2016} is:

\begin{equation}\label{eq:BN_orig}
\text{BN}\left(\mathbf{h};\gamma,\beta \right) = \beta + \gamma \odot \frac{ \mathbf{h} - \widehat{\mathbb{E}}[\mathbf{h}]}{\sqrt{\widehat{\text{Var}}[\mathbf{h}] + \epsilon}}
\end{equation}
where $\beta$ and $\gamma$ are the offset and scale, fixed at 0 and 0.1 in practice.
The BNLSTM above assumes the observations come from a single domain (Fig. \ref{fig:lstm_train_3}). We next propose to expand this to work on samples from two domains, $\mathcal{D}_1$, $\mathcal{D}_2$.  

Fig.~\ref{fig:lstm_train_4} introduces DDLSTM, proposed in this paper, which is designed to jointly learn temporal dependencies in two datasets, utilising the power of LSTMs in learning both short- and long-term dependencies.
We do this by first using concatenated label vectors. If $\mathcal{D}_1$ has $L_1$ labels and $\mathcal{D}_2$ has $L_2$ labels, then the one-hot label vector provided to the LSTM will be $L_1 + L_2$ dimensional.  Labels corresponding to $\mathcal{D}_1$ occupy entries $0$ to $L_1 - 1$, and labels corresponding to  $\mathcal{D}_2$ occupy entries $L_1$ to $L_1 + L_2 - 1$.
Another, more crucial, modification is to construct each batch with samples from $\mathcal{D}_1$ \emph{and} $\mathcal{D}_2$, so the BNLSTM can be jointly trained on both.  
The first $n_1$ samples in the batch are from $\mathcal{D}_1$, while the rest ($n_2 = N - n_1$) are from $\mathcal{D}_2$.

The standard BNLSTM is not suitable for dual domains, as the two domains are likely to have different means and variances.
In order to address this issue, a separate batch normalisation can be performed for samples from each domain. This would be sufficient but would, yet again, ignore any shared (cross-domain) temporal dependencies.
{We aim to learn cross-contamination between domains, when calculating batch statistics, as follows.}

\begin{figure}
\centering
\includegraphics[width=0.8\columnwidth]{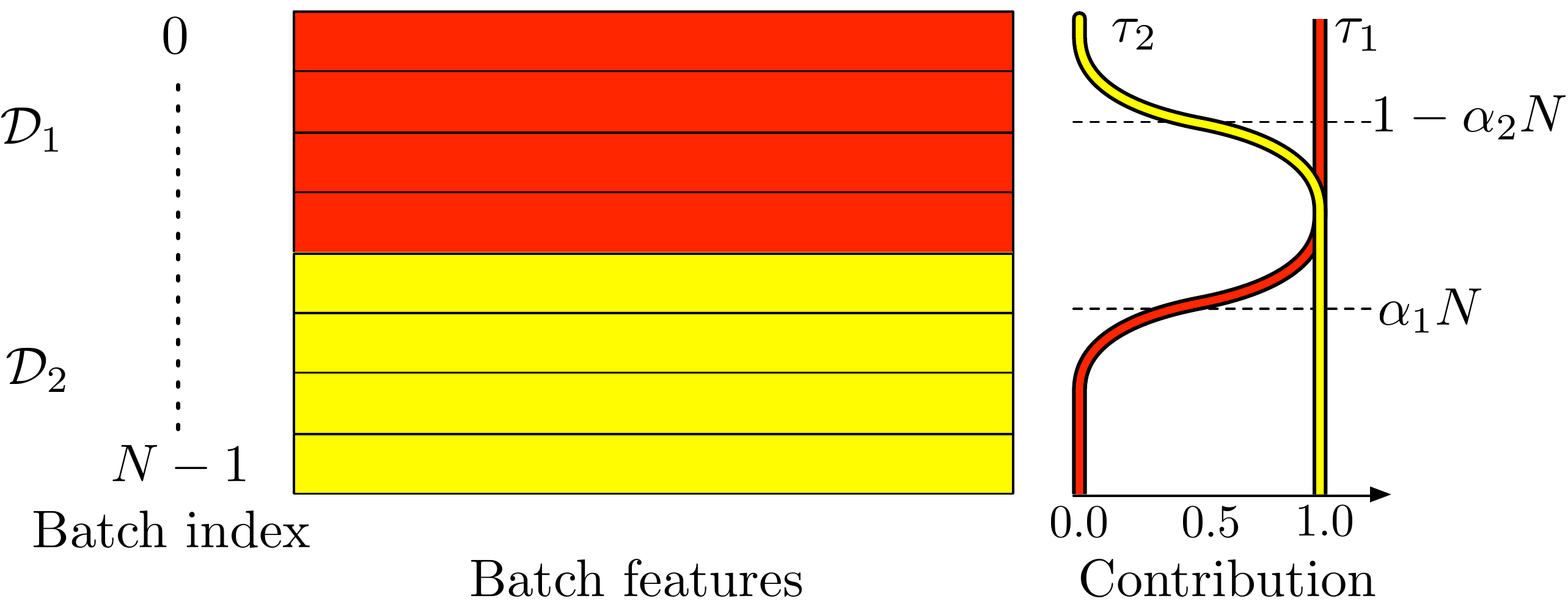}
\vspace*{-4pt}
\caption{Illustration of the contribution functions given in Equations \ref{eq:tau_1} (red) and \ref{eq:tau_2} (yellow).  The contribution of samples from both $\mathcal{D}_1$ and $\mathcal{D}_2$ to $\mathcal{D}_1$'s batch statistics is given by $\tau_1$, which is governed by the variable $\alpha_1$.  Similarly, $\alpha_2$ governs $\tau_2$, which gives the contribution of samples from both datasets to the batch statics of $\mathcal{D}_2$.}
\label{fig:contribution}
\vspace{-2mm}
\end{figure}

For each domain $\mathcal{D}_i$, we aim to learn {a corresponding cross-contamination factor} $\alpha_i$ 
which is used to determine the contributions of samples from the other domain to be included in the mean and variance calculations.  Each $\alpha_i$ is constrained such that $n_i \leq \alpha_i N \leq N$.  A higher $\alpha_i$ indicates that more cross-contamination occurs and vice versa.
Note that this cross contamination is required for accurate variance calculation -- if only means were required, then a weighted average of means for $\mathcal{D}_1$ and $\mathcal{D}_2$ could be used. The contribution function, $\tau_i$, then determines the contribution of the $j$'th sample in the batch for each domain, for a given parameter $\alpha_i$.
Each domain has its own contribution function (remember that samples from $\mathcal{D}_1$ appear first in each batch), defined as:
\vspace{-2mm}
\begin{equation}\label{eq:tau_1}
\tau_1(\alpha_1, j) = \frac{1-\text{tanh}(j - \alpha_1 N)}{2}
\end{equation}
\vspace{-2mm}
\begin{equation}\label{eq:tau_2}
\tau_2(\alpha_2, j) = \frac{1+\text{tanh}(j - \alpha_2 N)}{2}
\end{equation}
An illustration of this process is given in Fig \ref{fig:contribution}.

This can be used to redefine the batch normalisation function 
which learns from one domain (Equation \ref{eq:BN_orig}) with a dual-domain batch normalisation function, DDBN:
\begin{equation}\label{eq:DDBN}
\text{DDBN}\left(\mathbf{h};\gamma,\beta, \alpha_1, \alpha_2 \right) = \beta + \gamma \odot \frac{ \mathbf{h} - \widehat{\text{DD}\mathbb{E}}[\mathbf{h}, \alpha_1, \alpha_2]}{\sqrt{\widehat{\text{DD}\text{Var}}[\mathbf{h}, \alpha_1, \alpha_2] + \epsilon}}
\end{equation}
Instead of using standard expectation and variance calculations, DDBN relies on the contribution functions given in Equations \ref{eq:tau_1} and \ref{eq:tau_2} to give the expectation and variance for a weight $w$ at a specific timestep from each $\mathcal{D}_i$ as:
\begin{equation}\label{eq:expectation}
\text{DD}\mathbb{E}_i(w) = \frac{\sum_{j=1}^{N} w^j \tau_i(\alpha_i, j)}{\sum_{j=1}^{N}\tau_i(\alpha_i, j)}
\end{equation}
\begin{equation}\label{eq:variance}
\text{DD}\text{Var}_i(w) = \frac{\sum_{j=1}^{N} \left( w^j - \mathbb{E}_i(w) \right)^2 \tau_i(\alpha_i, j)}{\sum_{j=1}^{N}\tau_i(\alpha_i, j)}
\end{equation}
where $w^j$ denotes the value of $w$ corresponding to the $j$'th sample in the batch.

\begin{figure}
\centering
\subfloat[Joint training\label{fig:ddlstm_train_batch}]{\includegraphics[scale=0.19]{images/layout_ddlstm.pdf}} \hspace{3pt}
\subfloat[Test on $\mathcal{D}_1$ \label{fig:ddlstm_test_1}]{\includegraphics[scale=0.19]{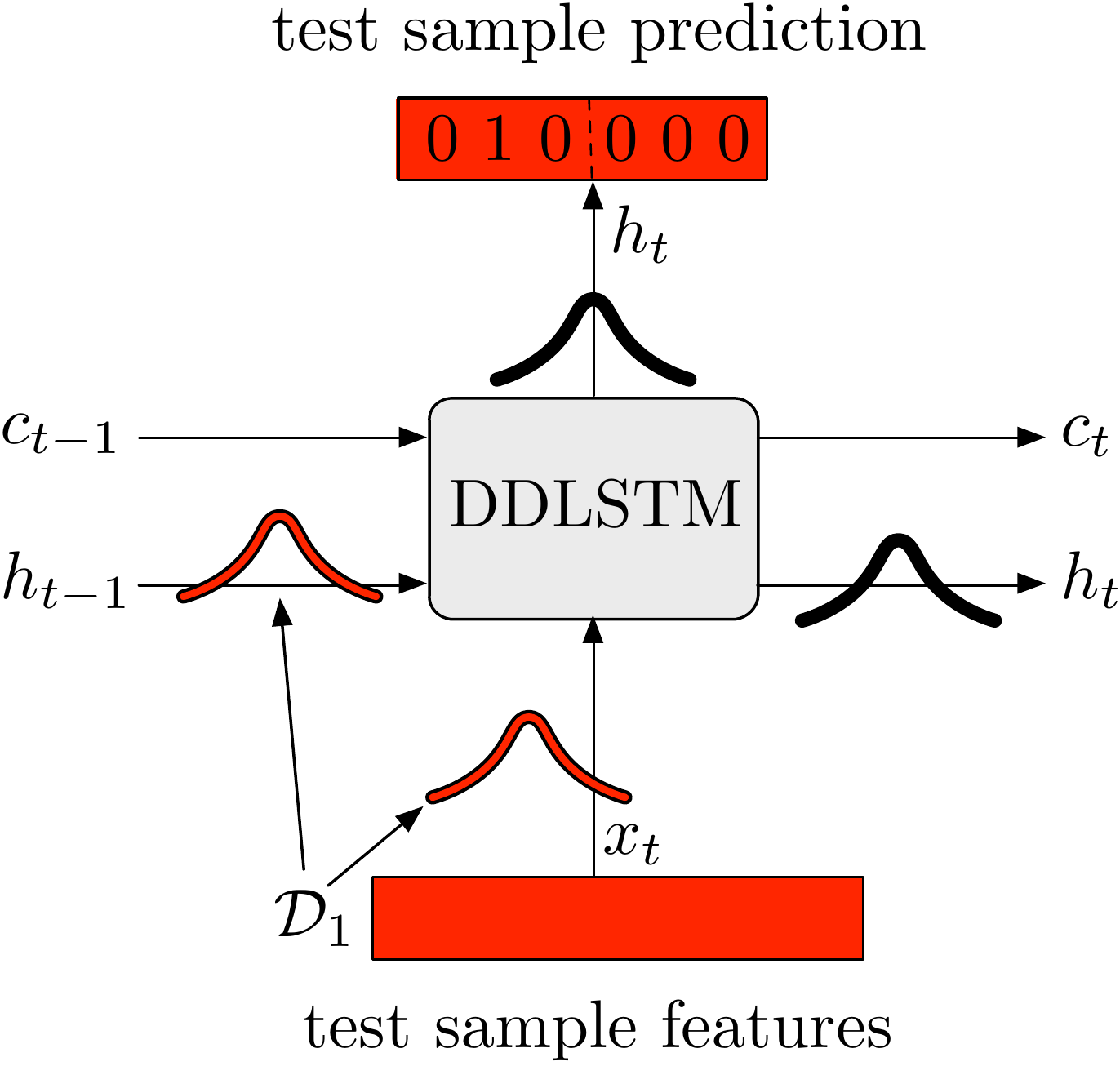}} \hspace{3pt}
\subfloat[Test on $\mathcal{D}_2$ \label{fig:ddlstm_test_2}]{\includegraphics[scale=0.19]{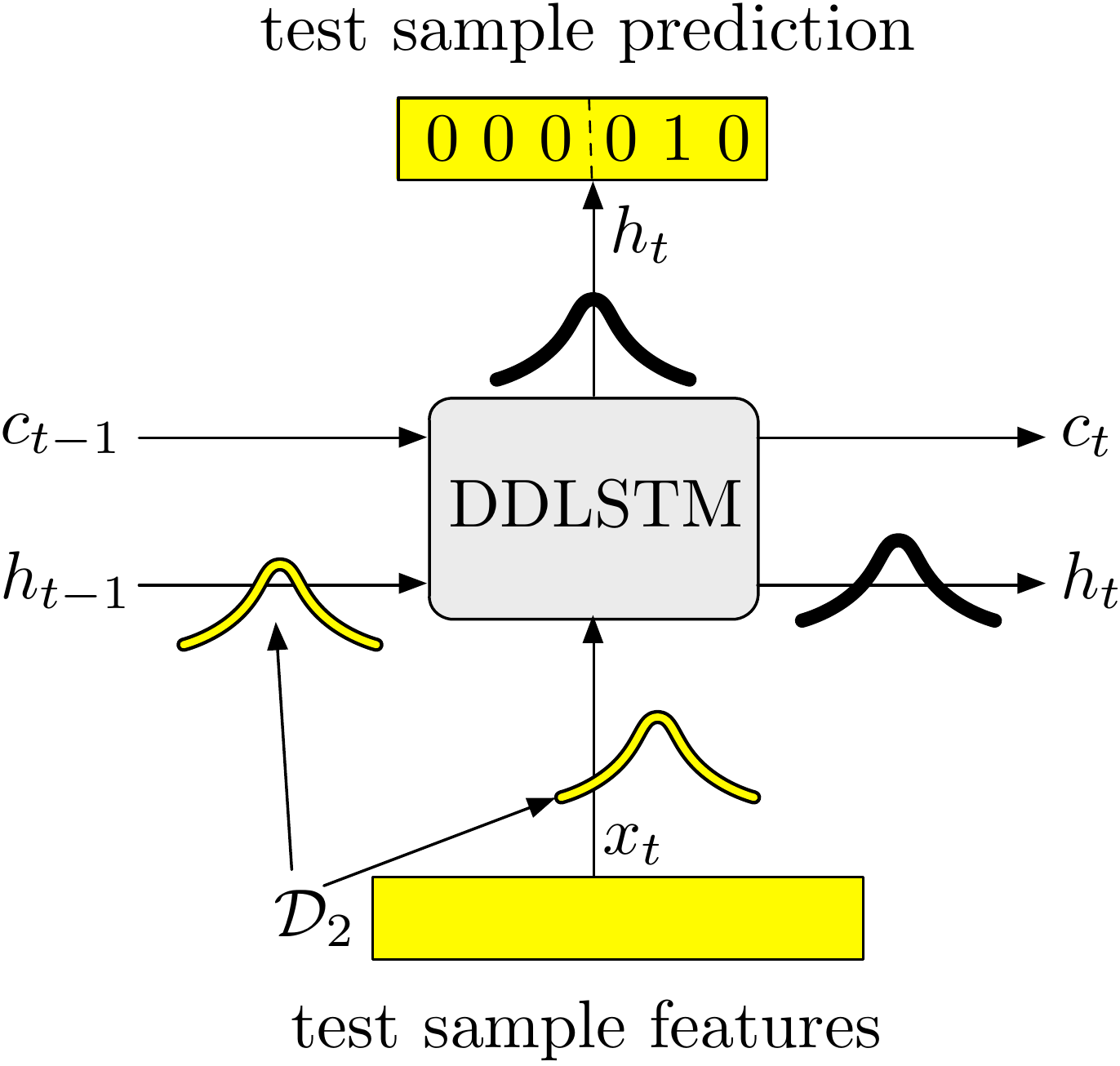}} 
\caption{DDLSTM joint training and testing on each dataset in turn.  During testing, a flag specifying which dataset the sample comes from is passed to the DDBN layers in order to use the correct batch statistics.}
\label{fig:train_test}
\vspace{-2mm}
\end{figure}

The main advantage of operating on the $\alpha$ values with $\text{tanh}$, rather than just selecting samples as in \cite{Carlucci2017}, is that it allows the whole process to be differentiable, and $\alpha$ values can be learned as part of the LSTM backpropogation-through-time process.
During training, $\text{DD}\mathbb{E}_i$ and $\text{DD}\text{Var}_i$ are estimated from the batch, and population wide estimates are updated for both domains as more batches are processed.  When testing an unseen sample, a flag is passed indicating which dataset the sample belongs to, and the dataset's population estimates are used for normalisation.  Fig. \ref{fig:train_test} shows this training and testing process.

\begin{figure*}[t]
\centering
\includegraphics[width=0.7\textwidth]{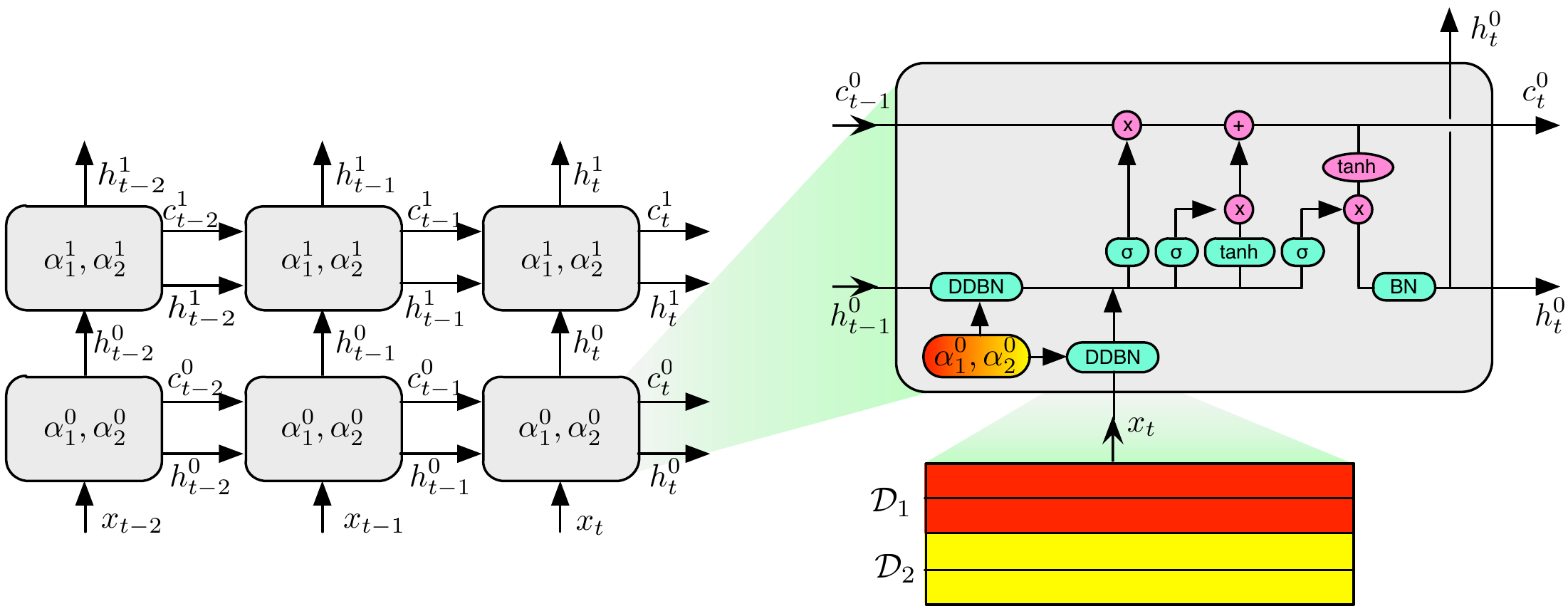}
\caption{Multi-layer DDLSTM architecture (left), and a DDLSTM cell (right). Note that $\alpha$s are shared, along with other LSTM parameters, between timesteps of the same level.  However, the dual-domain batch normalisation (which uses $\alpha$ values to determine the amount of cross-contamination) is calculated separately for each timestep.}
\label{fig:2x3}
\vspace{-2mm}
\end{figure*}

Note in Fig.~\ref{fig:lstm_train_4}, two DDBN functions are applied: input-to-hidden, and hidden-to-hidden. While a different set of $\alpha$ parameters could be used for each, we found that a single set of parameters gives better performance, 
presumably because there is similar dataset crossover at the hidden-to-hidden and input-to-hidden stages, and fewer parameters need to be learned.  Given these findings, we can define the DDLSTM, highlighting proposed differences in blue, as:
\begin{equation}\label{eq:ddlstm_f}
\left( \begin{array}{c}
\mathbf{\tilde f}_t \\
\mathbf{\tilde i}_t \\
\mathbf{\tilde o}_t \\
\mathbf{\tilde g}_t \\
\end{array} \right)
\begin{array}{ll}
 = & \text{\textcolor{blue}{DDBN}} \left( \mathbf{W}_h\mathbf{h}_{t-1};\gamma_h,\beta_h, \textcolor{blue}{\alpha_1, \alpha_2} \right) \\
  & + \text{\textcolor{blue}{DDBN}} \left( \mathbf{W}_x\mathbf{x}_{t};\gamma_x,\beta_x, \textcolor{blue}{\alpha_1, \alpha_2} \right)
 + \mathbf{b}
\end{array}
\end{equation}
with $\mathbf{c}_t$ and $\mathbf{h}_t$ defined as previously in Equations~\ref{eq:bnlstm_c} and~\ref{eq:bnlstm_h}.
DDBN was also trialled instead of the BN function in Equation \ref{eq:bnlstm_h}, and found to be be less effective (with regard to both the results and stability during training).  We hypothesise that this is because there is little point in effectively performing DDBN on the same data twice, and it would confuse the calculation of the final class probabilities.  

{Fig. \ref{fig:2x3} extends the proposed DDLSTM to two layers, and highlights where in the cell different forms of batch normalisation occur.}
Note {that} the parameters $\alpha$ are shared between timesteps, along with the rest of the LSTM cell, but the dual-domain batch normalisations are run individually for each timestep.  Whilst the number of samples from $\mathcal{D}_1$ ($n_1$) and $\mathcal{D}_2$ ($n_2$) in each batch can vary, in our experiments we set $n_1 = n_2 = N/2$.  If either $n_1$ or $n_2$ are zero (i.e. training data is only drawn from one domain), then the DDLSTM reduces to the BNLSTM \cite{Cooijmans2016}. 

{The Multi-layer DDLSTM can benefit from a gradual increase in cross-contamination, from lower to higher layers. This is based on the assumption that more shared information will be present in higher-layer representations than domain-specific lower-layers.
This was shown to be the case in CNNs~\cite{Carlucci2017} where cross-contamination increases as the network goes from deep to shallow.}
In Section \ref{ch:exp}, we test DDLSTM architectures with up to 10 layers, and show 3 layers give the best performance.

\begin{figure*}[t]
\centering
\subfloat[Breakfast]{\includegraphics[width=0.32\textwidth]{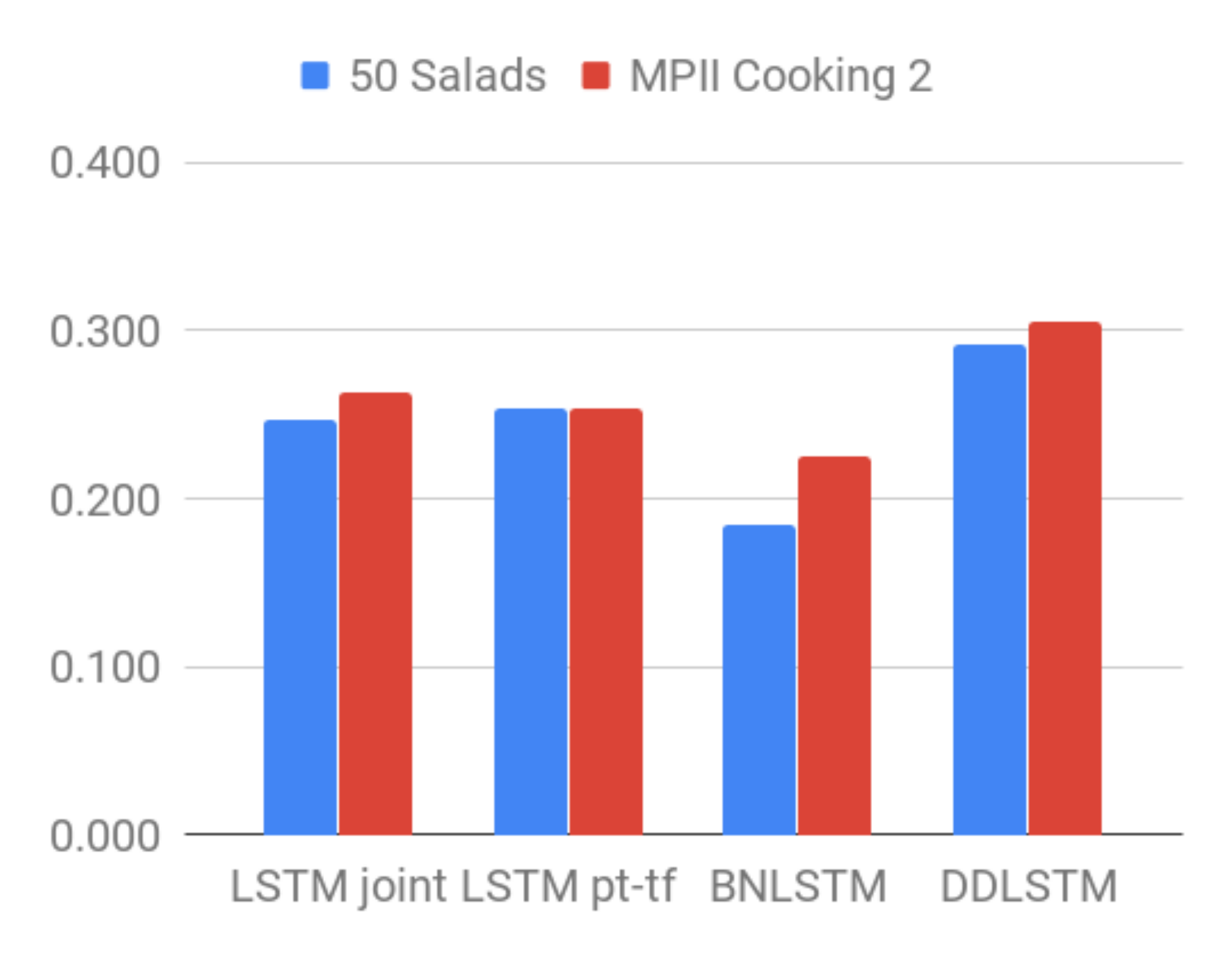}}
\subfloat[50 Salads]{\includegraphics[width=0.32\textwidth]{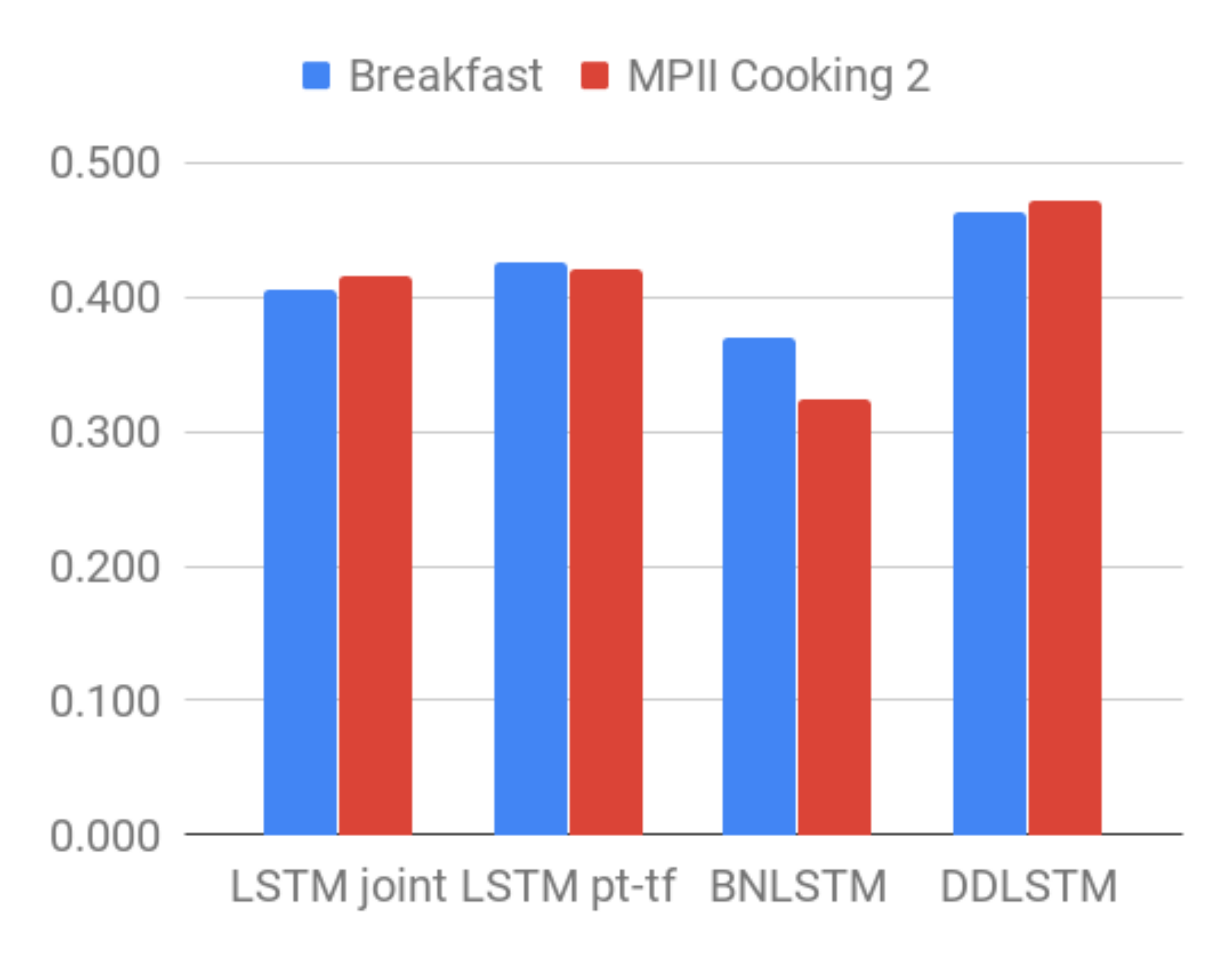}}
\subfloat[MPII Cooking 2]{\includegraphics[width=0.32\textwidth]{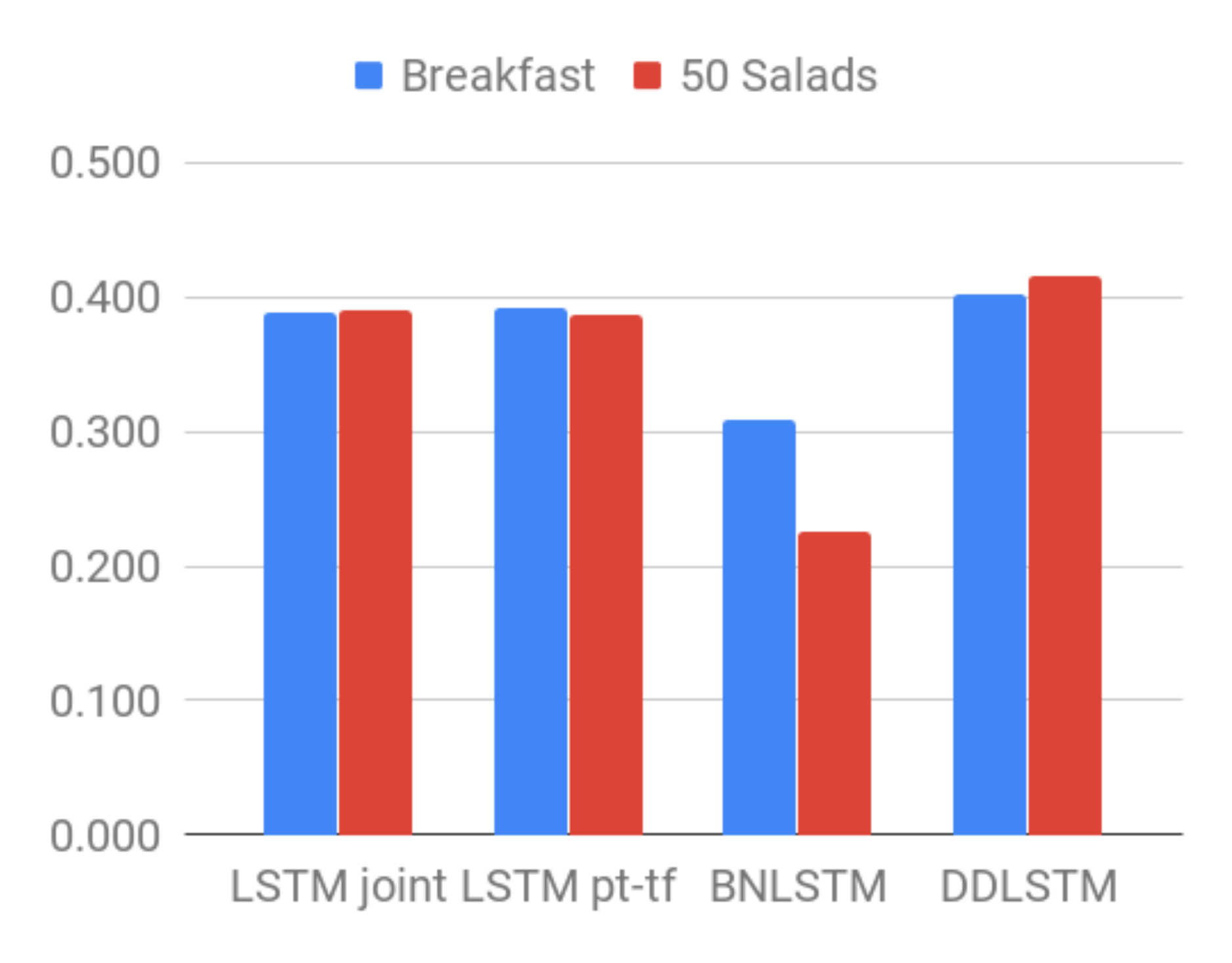}}
\vspace*{-8pt}
\caption{Visualised comparative results for each dataset. Colours indicate the second dataset used in DDLSTM in each case. For example, (a) shows the results for Breakfast trained with 50 Salads (blue) or trained with MPII Cooking (red).}
\label{fig:summary_results}
\end{figure*}

\section{Datasets}\label{ch:data}

\begin{figure*}[t]
\centering
\includegraphics[width=\textwidth]{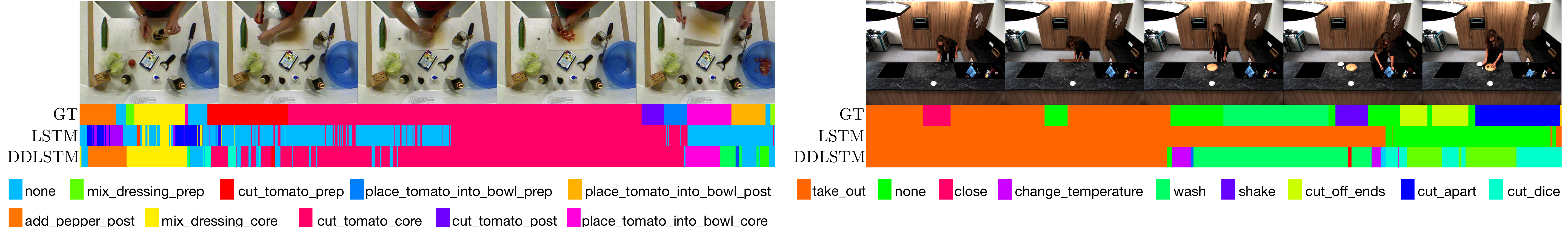}
\caption{A 1000 frame section from 50 Salads (left) and MPII cooking 2 (right).  GT shows the ground truth, LSTM indicates a standard LSTM fine-tuned, and DDLST is jointly trained on (Breakast and 50 Salads left; MPII and 50 Salads right).}
\vspace{-2mm}
\label{fig:qtv}
\end{figure*}

For this work, we use the three largest datasets of cooking-related activities with framewise action labels, which are all based around {activities} in the kitchen.  These are the Breakfast \cite{Kuehne2014}, 50 Salads \cite{Stein} and MPII Cooking~2~\cite{Rohrbach2012} datasets. There is very little label crossover as they are captured in different environments, with different viewpoints, participants and recipes.  However, we assume temporal dependences in the common tasks can be leveraged during a shared training process.
Some visual examples are given, along with qualitative results of some experiments, in Figure \ref{fig:qtv}.  For all datasets in this paper, 4 train/test splits are used, with 75\% training and 25\% testing. {All splits use leave-person-out, i.e. n}o participant appears in both training and testing sets from the same split.

Note that datasets such as UCF \cite{Soomro2012}, HMDB \cite{Kuehnea} and Kinetics \cite{Carreira} are not suitable here because they only contain a single action class per video sequence.  For online action classification, where each frame is classified as soon as it is seen, multiple actions are required for each video to ensure a robust evaluation.  Datasets which would fit this criteria include THUMOS \cite{Idrees2017} and ActivityNet \cite{Heilbron2015}, but they lack the task related-ness of Breakfast, 50 Salads and MPII Cooking, as we show in Section~\ref{sec:others}.

\noindent \textbf{Breakfast~\cite{Kuehne2014}:} \hspace{4pt}
The Breakfast dataset contains 433 sequences performed by 52 participants, containing 3078 actions across 50 classes (including a background class) in 18 different kitchens.  
All the sequences are one of 10 breakfast routines such as ``cooking scrambled eggs'' and ``making tea'', and no specific recipes are followed.  For the experiments in this paper, the lowest level action labels are used.  Examples include ``pour cereal'' and ``smear butter''.

\noindent \textbf{50 Salads~\cite{Stein}:} \hspace{4pt}
The 50 Salads dataset contains 50 videos, by 25 participants.  There are 52 of the lowest level action classes (including the background class which we have added), which gives a total of 2967 labelled actions.  
Example actions include ``cut tomato prep'', ``cut tomato core'', and ``cut tomato post''.  These prep- and post- labels are not found in the other two datasets.

\noindent \textbf{MPII Cooking 2 Salads~\cite{Rohrbach2012}:} \hspace{4pt}
The MPII Cooking 2 dataset contains 275 sequences from 30 participants in one kitchen.
It consists of 14105 actions across 88 classes (including the background class which we have added).  Example actions include ``shake'', ``spread'', and ``apply plaster''.

\begin{table}[t]
\centering
\resizebox{\linewidth}{!}{
\begin{tabular}{|l|l|l|l|l|l|}
\hline
D1        & D2      & Training & LSTM Type     & D1 Avg & D2 Avg \\ \hline
50 Salads    & MPII Cooking 2 & Single    & None & 41.1 & 38.3  \\
50 Salads    & MPII Cooking 2 & Joint    & LSTM & 41.6 & 39.0  \\
50 Salads    & MPII Cooking 2 & Joint    & BNLSTM       & 32.4 & 22.6  \\
50 Salads    & MPII Cooking 2 & D1,D2    & LSTM & 04.4 & 38.7  \\
50 Salads    & MPII Cooking 2 & D2,D1    & LSTM & 43.0 & 00.0  \\
50 Salads    & MPII Cooking 2 & Joint    & DDLSTM   & \bf{47.1} & \bf{41.5}  \\ \hline
Breakfast & 50 Salads  & Single    & None & 24.5 & 41.1  \\
Breakfast & 50 Salads  & Joint    & LSTM & 24.7 & 40.5  \\
Breakfast & 50 Salads  & Joint    & BNLSTM       & 18.4 & 37.0  \\
Breakfast & 50 Salads  & D1,D2    & LSTM & 00.0 & 42.5  \\
Breakfast & 50 Salads  & D2,D1    & LSTM & 25.4 & 08.5  \\
Breakfast & 50 Salads  & Joint    & DDLSTM   & \bf{29.1} & \bf{46.3}  \\ \hline
Breakfast & MPII Cooking 2 & Single    & None & 24.5 & 38.3  \\
Breakfast & MPII Cooking 2 & Joint    & LSTM & 26.3 & 38.8  \\
Breakfast & MPII Cooking 2 & Joint    & BNLSTM       & 22.5 & 30.9  \\
Breakfast & MPII Cooking 2 & D1,D2    & LSTM & 00.0 & 39.1  \\
Breakfast & MPII Cooking 2 & D2,D1    & LSTM & 25.3 & 01.0  \\
Breakfast & MPII Cooking 2 & Joint    & DDLSTM   & \bf{30.5} & \bf{40.1} \\ \hline
\end{tabular}
}
\caption{Average results, over splits, on pairs of the three datasets.  A comparison of the Dual-Domain LSTM is compared to BNLSTM and standard LSTM using different training approaches (joint training, pre-training on D1 and fine-tuning on D2 and vice versa). None: frame-level classification without any temporal modelling. 
}
\label{tab:summary_results}
\vspace{-2mm}
\end{table}

\section{Experiments}\label{ch:exp}

In this section we detail frame-based feature extraction, and provide comparative analysis against other LSTM architectures. 
In all experiments, frame-based classification accuracy is reported. 
\textbf{Implementation Details:} For each LSTM type, 128 hidden units are used per cell. We use a batch size of 128 with a 50/50 split between datasets, so each batch contains 64 sequences from each of the two datasets being evaluated.  A learning rate of 0.01 is used for 50,000 iterations, and all $\alpha$ values are initialised to 0.75. Other values between 0.5 and 1 were trialled, but made little difference.

\subsection{Online frame-based Results}

We are particularly interested in online action recognition, that is the ability to recognise actions given current and past observations, without an insight into future frames.
The input to each of the tested LSTM architectures are frame-level features extracted from a CNN.
For each split (from each dataset), Inception V2 \cite{Szegedy2014a} (initialised with HMDB51 weights \cite{Kuehnea}) is trained using the training split to classify frames individually.  
This model is then used to extract features for test images.
The activations from the last layer of the network (i.e. the logits) are extracted to use as features.  Baselines for single datasets are given in Table \ref{tab:summary_results}.

\begin{figure*}[t]
\centering
\begin{minipage}[h]{0.6\textwidth}
\includegraphics[width=1\textwidth]{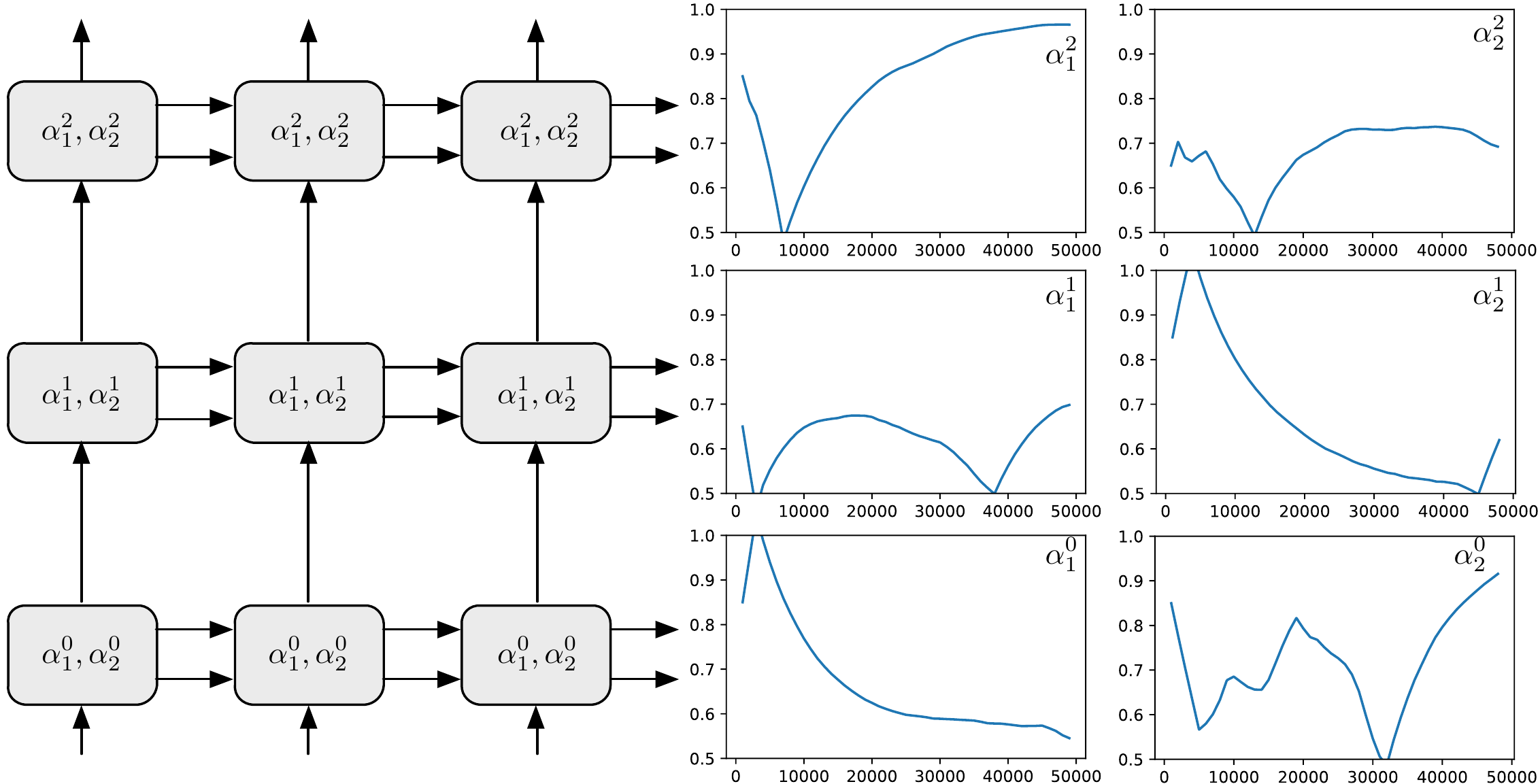}
\end{minipage}
\hspace{0.03\textwidth}
\begin{minipage}[h]{0.33\textwidth}
$\alpha_1^2$
\includegraphics[width=0.8\linewidth]{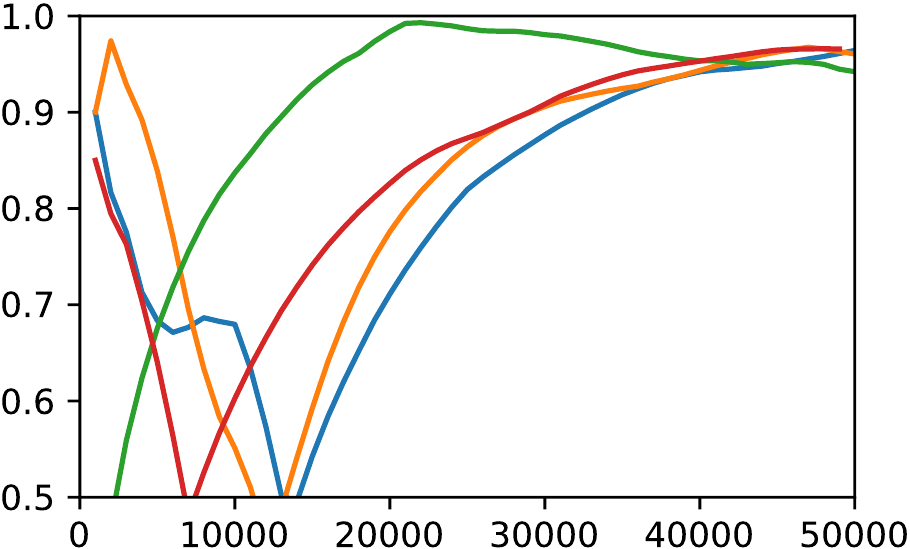}\\
$\alpha_2^2$
\includegraphics[width=0.8\linewidth]{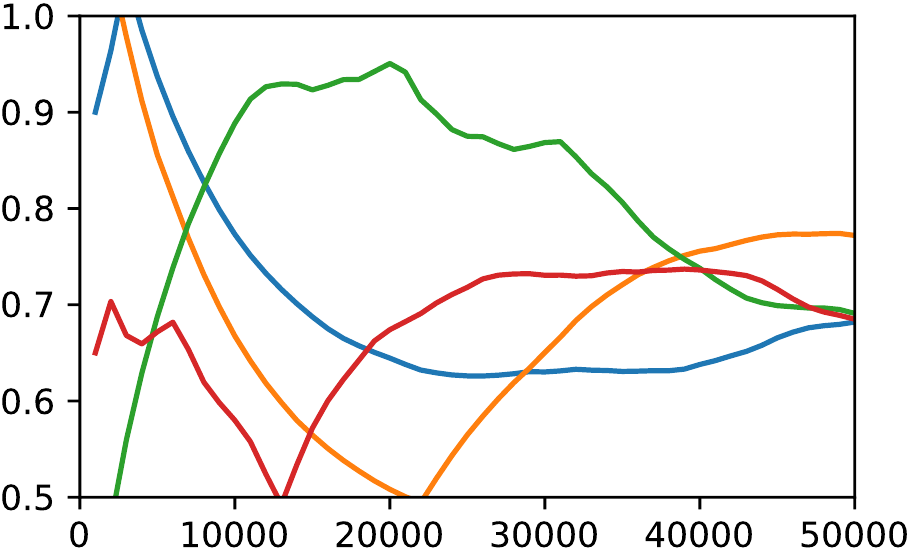}
\caption{Examples of $\alpha$ progress during training.  The graphs on the right show how $\alpha$'s perform with different initialisations.}
\label{fig:alpha_inits}
\end{minipage}
\end{figure*}

It has been shown that directly predicting the next action boundary provides better performance on frame-based classification problems \cite{Shen2015}.  Following the method for prediction in this work, we first train an LSTM architecture (1 layer deep, history size 200 frames), where the loss for a sequence is the KL divergence between Gaussians at the next action boundary and the current boundary prediction for every frame.
This boundary prediction and the original features are then fed as input to the various LSTM architectures (3 layers deep with a history of 200 frames), trained using a softmax loss.
Concatenated label vectors, as illustrated earlier in Fig. \ref{fig:comparison}, are used for all experiments.
We evaluate frame-based action recognition on the three datasets introduced in Section \ref{ch:data}.  

The following LSTM architectures and training protocols are compared:
\begin{itemize}[leftmargin=*]
\vspace*{-6pt}
\item LSTM jointly trained on two datasets.
\vspace*{-6pt}
\item LSTM pre-trained on one and fine-tuned on the other.
\vspace*{-6pt}
\item BNLSTM \cite{Cooijmans2016} jointly trained on two datasets.
\vspace*{-6pt}
\item DDLSTM jointly trained on two datasets.
\vspace*{-6pt}
\end{itemize}
Table \ref{tab:summary_results} gives the results for these experiments, averaged over all four splits.
A corresponding visualisation for these averages can be seen in Fig. \ref{fig:summary_results}.  It shows that the DDLSTM outperforms the jointly trained and fine-tuned LSTMs as well as the jointly trained BNLSTM across all pairs of datasets.  Compared to the next best performing method for each dataset pair, where the second best is a different method for each case, there are increases of 5.1\% and 2.5\% for 50 Salads and MPII Cooking~2, 3.7\% and 3.8\% for Breakfast and 50 Salads and 4.2\% and 1.0\% for Breakfast and MPII Cooking~2. 
It is expected that MPII Cooking~2, being the largest dataset, would benefit less from cross-dataset training than the two smaller datasets.
Qualitative results are shown in Fig.~\ref{fig:qtv}, comparing DDLSTM to the second best performing LSTM architecture in each case.

\subsection{Discussion}
Table \ref{tab:layers} gives an evaluation of how the number of levels (or depth) of the DDLSTM architecture affects performance.  In general, there is a marginal improvement as the number of levels increases up to 3 with a drop off afterwards, although we observed one case where there was a slight drop (MPII Cooking 2 when trained with Breakfast).

\begin{table}[t]
\centering
\resizebox{1\textwidth}{!}{
\begin{tabular}{|l|l|l|l|l|}
\hline
D1        & D2      & LSTM Layers & D1 Avg & D2 Avg \\ \hline
50 Salads    & MPII Cooking 2 & 1 & 45.7 &	41.3 \\
50 Salads    & MPII Cooking 2 & 2 & 46.6 &	41.0 \\
50 Salads    & MPII Cooking 2 & 3 & \bf{47.1} &	\bf{41.5} \\ 
50 Salads    & MPII Cooking 2 & 4 & 46.1 &	40.6 \\ 
50 Salads    & MPII Cooking 2 & 5 & 41.0 &	40.3 \\ 
50 Salads    & MPII Cooking 2 & 10* & 39.7 &	38.0 \\ 
\hline
Breakfast & 50 Salads  & 1        & 28.4 &	45.0 \\
Breakfast & 50 Salads  & 2        & 29.0 &	45.2 \\
Breakfast & 50 Salads  & 3        & \bf{29.1} &	\bf{46.3} \\ 
Breakfast & 50 Salads  & 4        & 28.6 &	44.3 \\ 
Breakfast & 50 Salads  & 5        & 29.0 &	46.0 \\ 
Breakfast & 50 Salads  & 10*        & 25.5 & 40.7	 \\ 
\hline
Breakfast & MPII Cooking 2 & 1    & 29.2 &	\bf{40.8} \\
Breakfast & MPII Cooking 2 & 2    & 29.9 &	40.1 \\
Breakfast & MPII Cooking 2 & 3    & \bf{30.5} &	40.1 \\ 
Breakfast & MPII Cooking 2 & 4    & \bf{30.5} &	38.8 \\ 
Breakfast & MPII Cooking 2 & 5    & 30.3 &	38.4 \\ 
Breakfast & MPII Cooking 2 & 10*    & 28.1 & 35.6	 \\ 
\hline
\end{tabular}
}
\caption{Average accuracies over all dataset splits, comparing DDLSTM architecture with different depths 1-10. {\footnotesize *Depth 10 used a history of 100 due to memory constraints}.}
\label{tab:layers}
\end{table}

Fig. \ref{fig:alpha_inits} shows an example of how the $\alpha$ values, which control the dual-domain cross-contamination, change during training.  In \cite{Carlucci2017}, automatic domain alignment layers were shown to use more cross-contamination for high-level layers than low-level layers.  We sometimes observed analogous behaviour, such as $\alpha_1^0$, $\alpha_1^1$ and $\alpha_1^2$. Here, $\alpha_1^0 \approx 0.5$ indicates no cross-contamination, while $\alpha_1^2 \approx 1$ indicates total cross-contamination.
However, other behaviours were also seen, such as $\alpha_2^0$, $\alpha_2^1$ and $\alpha_2^2$.  One possible explanation is that the DDLSTMs are already being fed high-level features, so there is a less drastic deep-to-shallow progression in terms of what these features represent.  This could also be a reason for there being a small, rather than large, improvement when increasing the depth of the DDLSTM network (Table \ref{tab:layers}).

We noted earlier that the method is robust to initialisations of $\alpha$ values.  
Fig. \ref{fig:alpha_inits} also shows how different initialisations converge to similar $\alpha$ values after training. 
We note that $\alpha$ values from multiple runs are unlikely to be identical, given that batches contain random sample orderings.

\subsection{Comparison to Published Results}

Very few works have attempted online frame-level accuracy on these datasets, with most works focusing on offline classification~\cite{Huang2016,Singh2016,Lea,Jiao2018,Song2018}. To compare to these, we also evaluate DDLSTM with look-ahead, i.e. allowing it to see half the training history size into the future when classifying the current frame. 

In Table~\ref{tab:offline_breakfast}, we report results on Breakfast, showing our method outperforms published results operating online - note that~\cite{DeGeest18} only provides results on a single split in the dataset. 
We use ``offline multi-pass'' to refer to methods that iteratively optimise the classification of frames, by multi-pass segmentation of the whole video. ``Offline single pass'' methods have access to future frames, but only as a single pass, e.g.~bi-directional LSTM. ``Online'' methods classify each frame as soon as it is seen, with no access to future frames, e.g.~uni-directional LSTM.
We do not expect our single-pass results to outperform multi-pass offline evaluations, but provide these results for completion. 

\begin{table}[]
\centering
 \resizebox{1\linewidth}{!}{
\begin{tabular}{|l|l|l|}
\hline
Method      &Mode  & Accuracy \\ \hline
\cite{Kuehne2016} &offline multi-pass   & 56.3   \\
\cite{Richard18} &offline multi-pass  &43.0\\ 
DDLSTM (look-ahead) w/f \cite{Kuehne2016} &offline single-pass & 26.4 \\
DDLSTM (look ahead) &offline single-pass & 32.6 \\
\hline
\cite{Richard18} &online &27.2\\
\cite{DeGeest18} (only evaluated on 1/4 splits) &online &32.6\\ \cite{Perrett} &online &28.5 \\
DDLSTM w/f \cite{Kuehne2016} &online & 23.8\\
DDLSTM &online & 29.1 \\
\hline
\end{tabular}
 }
 \caption{Comparative analysis on the Breakfast dataset, using our features as well as with features (w/f) from \cite{Kuehne2016}. The DDLSTM uses the public features from 50 salads~\cite{Lea} (used in Table~\ref{tab:offline_salads_mid}) as its second domain.} \label{tab:offline_breakfast}
\end{table}

In Table~\ref{tab:offline_salads_mid}, we compare to published offline results on 50~Salads, using publicly available features from~\cite{Lea}. These use mid-level classes (17 classes plus a background class), which is significantly easier than what we report in Table~\ref{tab:summary_results} where we use all 52 lowest-level classes. Our online results and results with look ahead are only slightly worse than others evaluated offline. We have not found any published results for online performance on 50 salads mid-level, or any other results using publicly available features.

\begin{table}[]
\centering
 \resizebox{1\linewidth}{!}{
\begin{tabular}{|l|l|l|}
\hline
Method    &Mode    & Accuracy \\ \hline
Bi-LSTM (1 layer) \cite{Lea} &offline single-pass   & 55.7   \\
ED-TCN \cite{Lea} &offline single-pass   & 64.7   \\ 
DDLSTM (look-ahead) w/f \cite{Lea} &offline single-pass & 60.9 \\
\hline
LSTM w/f \cite{Lea} &online & 57.6 \\
DDLSTM w/f \cite{Lea} &online & 59.1\\
\hline
\end{tabular}
 }
 \caption{Comparative analysis on the 50 salads dataset, using mid-level classes with features (w/f) from \cite{Lea}. The DDLSTM uses the public features from Breakfast~\cite{Kuehne2016} as its second domain (used in  Table \ref{tab:offline_breakfast}).}

\label{tab:offline_salads_mid}
\end{table}

\subsection{Effect of Related Domain Adaptation}
\label{sec:others}
To investigate how much of the DDLSTM's improvement comes from exploiting related temporal information, we evaluated whether 50 Salads (split 0) benefits equally from three large-scale datasets with various levels of domain relatedness. We test on THUMOS \cite{Idrees2017} (features from \cite{gao2017cascaded}), ActivityNet \cite{Heilbron2015} (features from \cite{Heilbron2015}) and EPIC Kitchens \cite{Damen2018EPICKITCHENS} (ImageNet ResNet 50 features). Of these, only EPIC presents a related `kitchen' domain. THUMOS and ActivityNet capture actions unrelated to the kitchen domain and contain very few actions per sequence. 

We report results in Table~\ref{tab:other}. Only by fine-tuning, results show that the related dataset EPIC provides best performance.
We observe no benefit to in jointly training using DDLSTM with THUMOS or ActivityNet. We however observe clear improvements when jointly training with EPIC Kitchens, for both EPIC (by 1.6\%) and 50 Salads (by 4\%). We conclude that 1) DDLSTM is particularly suited for related domains, and that 2) a higher increase in accuracy is expected for smaller datasets as these leverage missing temporal knowledge from the larger dataset.

\begin{table}[t]
\centering
\resizebox{\linewidth}{!}{
\begin{tabular}{|l|l|l|l|l|l|}
\hline
D1        & D2      & Training & LSTM Type     & D1 Acc & D2 Acc \\ \hline
ActivityNet    & 50 Salads & Pt/ft    & LSTM     & 44.4 & 42.1  \\
ActivityNet    & 50 Salads & Joint    & DDLSTM   & 44.3 & 42.2  \\ \hline
Thumos    & 50 Salads & Pt/ft    & LSTM     & 65.9 & 42.0  \\
Thumos    & 50 Salads & Joint    & DDLSTM   & 66.1 & 42.3  \\ \hline
EPIC    & 50 Salads & Pt/ft    & LSTM     & 31.5 & \textbf{44.9}  \\
EPIC    & 50 Salads & Joint    & DDLSTM   & 33.1 & \textbf{48.9}  \\ \hline
\end{tabular}
}
\caption{Classification accuracy when learning 50 salads (split 0) from larger datasets. Pt/ft refers to results on D1 when pre-training on D2 and fine tuning on D1 and vice versa.  Figure shows that the large dataset with related domain (EPIC) performs better for pre-training, and shows larger improvement using DDLSTM.}
\label{tab:other}
\vspace{-2mm}
\end{table}

\section{Conclusion and Future Work}\label{ch:conclusion}

In this paper, the Dual-Domain LSTM (DDLSTM) was introduced, which is capable of learning temporal information from two domains at the same time.  Given batches consisting of samples from both domains, DDLSTM applies a dual-domain batch normalisation on both input-to-hidden and hidden-to-hidden LSTM weights.  This calculates separate batch statistics for each domain, but learns a parameter which determines how much cross-contamination between domains should be included in a fully differentiable manner.
The learnt parameters are shared across timesteps, but the batch normalisation calculation is performed on data at each timestep separately.
We evaluated the DDLSTM architecture on online action recognition, using three cooking datasets with multiple actions per video and framewise labels. DDLSTM was found to outperform (by a 3.5\% average across all datasets) the standard LSTM (both jointly trained, and pre-trained and fine-tuned) and the batch-normalised LSTM upon which it builds.

This paper presents a number of opportunities (A-D) for future investigation.  A) Learning from more than two related datasets/domains at the same time.  This would require modifications to the contribution functions (Equations~\ref{eq:tau_1} and~\ref{eq:tau_2}) and an increase in the number of $\alpha$s to $d(d-1)l$, where $d$ and $l$ are the number of datasets and LSTM layers.  B) Automatically adjusting the makeup of each batch could provide performance improvements, in a similar fashion to the way $\alpha$ values are leaned for cross-contamination. C) Using an attention mechanism \cite{Qin2017} to determine which items within a batch are most useful for cross-contamination.  D)~Incorporating frame-based domain adaptation methods into the feature extractor as well as the LSTM.

\noindent \textbf{Acknowledgement:} Research supported by EPSRC LOCATE (EP/N033779/1) and uses publicly available data.

\bibliographystyle{ieee}
\bibliography{library}
\end{document}